\documentclass[10pt,twocolumn,letterpaper]{article}

\usepackage{cvpr}              

\usepackage{graphicx}
\usepackage{amsmath}
\usepackage{amssymb}
\usepackage{booktabs}
\usepackage{dsfont}
\usepackage[hang]{footmisc}
\usepackage[pagebackref,breaklinks,colorlinks]{hyperref}

\newcommand\blfootnote[1]{%
  \begingroup
  \renewcommand\thefootnote{}\footnote{#1}%
  \addtocounter{footnote}{-1}%
  \endgroup
}
\usepackage[capitalize]{cleveref}
\crefname{section}{Sec.}{Secs.}
\Crefname{section}{Section}{Sections}
\Crefname{table}{Table}{Tables}
\crefname{table}{Tab.}{Tabs.}

\def\cvprPaperID{*****} 
\def\confName{CVPR}
\def\confYear{2022}

\begin{document}

\title{The Runner-up Solution for YouTube-VIS Long Video Challenge 2022}


\author
{
Junfeng Wu$^{1*}$
~~~
Yi Jiang$^{2}$
~~~
Qihao Liu$^{3*}$
~~~
Xiang Bai$^{1}$
~~~
Song Bai$^{2}$
\\[0.2cm]
${^1}$Huazhong University of Science and Technology ~~~
\\
${^2}$ByteDance Inc.~~~
${^3}$Johns Hopkins University ~~~
}

\maketitle
\blfootnote{
$*$ Work done during an internship at ByteDance. \\
}

\begin{abstract}

This technical report describes our 2nd-place solution for the ECCV 2022 YouTube-VIS Long Video Challenge.
We adopt the previously proposed online video instance segmentation method IDOL for this challenge.
In addition, we use pseudo labels to further help contrastive learning, so as to obtain more temporally consistent instance embedding to improve tracking performance between frames.
The proposed method obtains 40.2 AP on the YouTube-VIS 2022 long video dataset and was ranked second place in this challenge.
We hope our simple and effective method could benefit further research.
\end{abstract}

\section{Introduction}
\label{sec:intro}

Video instance segmentation aims at detecting, segmenting, and tracking object instances simultaneously in a given video. It has attracted considerable attention since first defined~\cite{MaskTrackRCNN} in 2019 due to the huge challenge and the wide applications in video understanding, video editing, autonomous driving, augmented reality, etc. Current VIS methods can be categorized as online or offline methods. Online methods~\cite{MaskTrackRCNN,sipmask,compfeat,STMask,CrossVIS,PCAN,IDOL} take as input a video frame by frame, detecting and segmenting objects per frame while tracking instances and optimizing results across frames. Offline methods~\cite{MaskProp,ProposeReduce,STEmSEG,VisTR,IFC,seqformer}, in contrast, take the whole video as input and generate the instance sequence of the entire video with a single step.

In this challenge, the videos are longer with the lower sampling rate, which masks algorithm easily lose targets ID due to the large movement of the objects and accumulate errors during the longer tracking process.
In addition, due to the extremely low sampling rate, the appearance similarity of objects between adjacent frames are smaller. 
All these characteristics degrade the performance of previous algorithms significantly and make this dataset very challenging.

To handle longer videos, as well as perform more robust tracking on low sample rate video frames, we use IDOL~\cite{IDOL} as our baseline algorithm.
IDOL is a online video instance segmentation method based on contrastive learning, which is able to ensure, in the embedding space, the similarity of the same instance across frames and the difference of different instances in all frames, even for instances that belong to the same category and have very similar appearances. It provides more discriminative instance embeddings with better temporal consistency, which guarantees more accurate association results.
To improve the temporal consistency of instance embeddings between low sampling rate frames, we propose to pre-train the model on COCO~\cite{coco} with pseudo-labels, which further improves the performance of IDOL.

Our method achieves the second place in the ECCV 2022 YouTube-VIS Long Video Challenge, with the score 53.6 AP on the public validation set, and 40.2 AP on the private test set. 
We believe the simplicity and effectiveness of our method shall benefit further research.

\begin{figure*}[tb]
\centering
\includegraphics[width=1\linewidth]{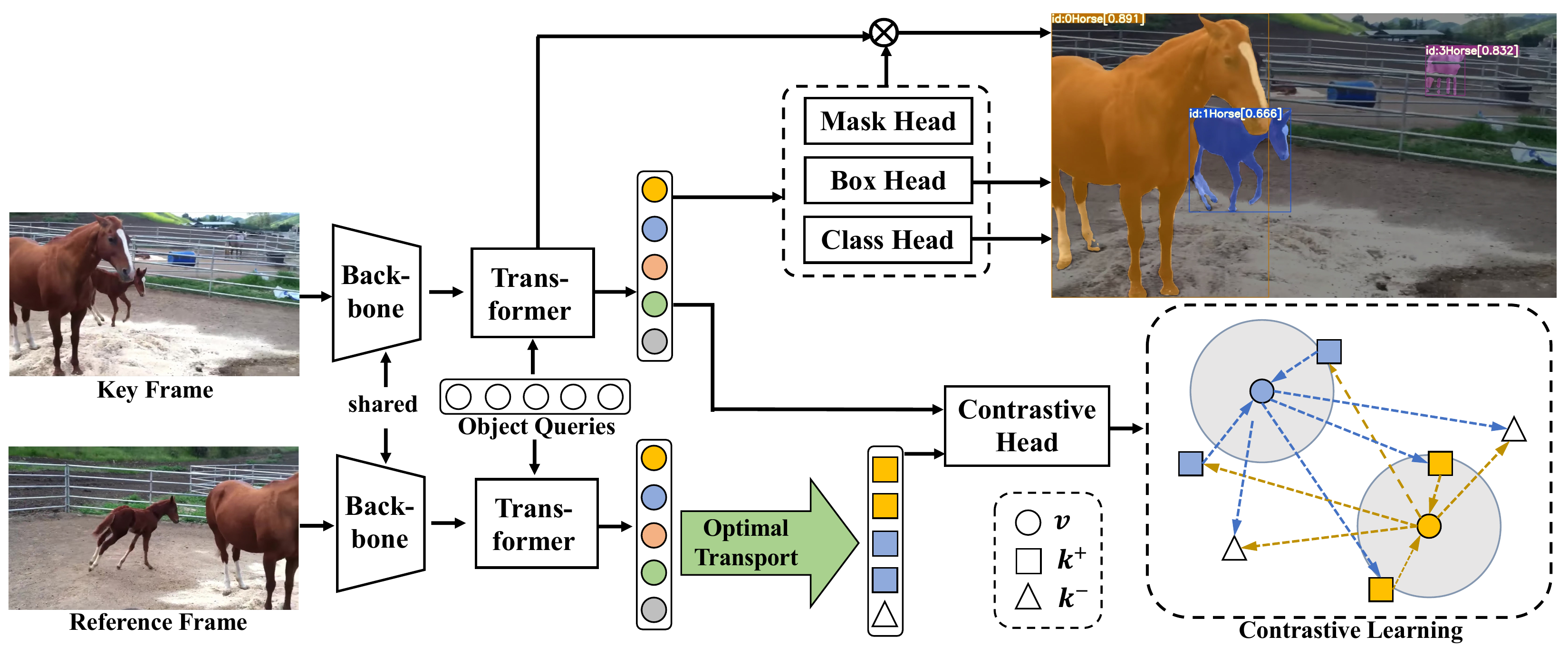}
\caption{The training pipeline of IDOL. 
	Given a key frame and a reference frame, the shared-weight backbone and transformer predict the instance embeddings on them respectively. The embeddings on the key frame are used to predict masks, boxes, and categories, while the embeddings on the reference frame are selected as positive and negative embeddings for contrastive learning.
	}
\label{fig:IDOL}
\end{figure*}

\section{Method}


\subsection{Video Instance Segmentation}
\label{sec:InstSeg}
We take IDOL~\cite{IDOL} as our baseline method.
Following most previous online VIS models~\cite{MaskTrackRCNN,CrossVIS} which utilize additional association head upon on instance segmentation models~\cite{MaskRCNN,CondInst}, IDOL takes DeformableDETR~\cite{deformableDETR} with dynamic mask head~\cite{CondInst} as its instance segmentation pipeline.

Given an input frame $x\in R^{3\times H\times W}$ of a video, a CNN backbone extracts multi-scale feature maps. The Deformable DETR takes the feature maps with additional fixed positional encodings~\cite{detr} and $N$ learnable object queries as input. The object queries are first transformed into output embeddings $ E \in R^{N\times C} $ by the transformer decoder. After that, they are decoded into box coordinates, class labels, and instance masks following SeqFormer~\cite{seqformer} as shown in Fig~\ref{fig:IDOL}.
Then it calculate pair-wise matching cost which takes into account both the class prediction and the similarity of predicted and ground truth boxes.

An extra light-weighted FFN is employed as a contrastive head to decode the contrastive embeddings from the output embeddings. Given a key frame for instance segmentation training, a reference frame from the temporal neighborhood is selected for contrastive learning. For each instance in the key frame, their output embedding with the lowest cost are decoded to the contrastive embedding $\textbf{v}$. 
If the same instance appears on the reference frame, 
we select positive and negative samples for it according to the cost with predictions.
The contrastive loss function for a positive pair of examples is defined as follows:
\vspace{-2mm}
\begin{equation} \label{eq:lossembed2}
 \mathcal{L}_{embed} = \log [1+\sum_{\textbf{k}^+}\sum_{\textbf{k}^-}\exp(\textbf{v} \cdot \textbf{k}^-  - \textbf{v} \cdot \textbf{k}^+) ].
 \vspace{-2mm}
\end{equation}
where $\textbf{k}^+$ \text{and} $\textbf{k}^-$ are positive and negative feature embeddings from the reference frame, respectively.
Finally, the whole model is optimized with a multi-task loss function
\vspace{-2mm}
\begin{equation} \label{eq:sumloss}
\mathcal{L} = \mathcal{L}_{cls}+\lambda_1\mathcal{L}_{box}+ \lambda_1\mathcal{L}_{mask}+\lambda_2\mathcal{L}_{embed},
\end{equation}

Given a test video, we initialize an empty memory bank for it and perform instance segmentation on each frame sequentially in an online scheme. 
Assume there are $N$ instances predicted by the model with $N$ contrastive embeddings, and $M$ instances in the memory bank. We compute similarity score $f$ between predicted instance $i$ and memory instance $j$, and search for the best assignment for instance $i$ by: 
\begin{equation} \label{eq:argmax}
\hat{j} = \arg\max \textbf{f}(i,j), \forall j \in\{1,2,...,M\}.
\end{equation}
If $\textbf{f}(i,\hat{j})>0.5$, we assign the instance $i$ on current frame to the memory instance $\hat{j}$. For the prediction without an assignment but has a high class score, we start a new instance ID in the memory bank.

\begin{figure}[tb]
\centering
\includegraphics[width=1\linewidth]{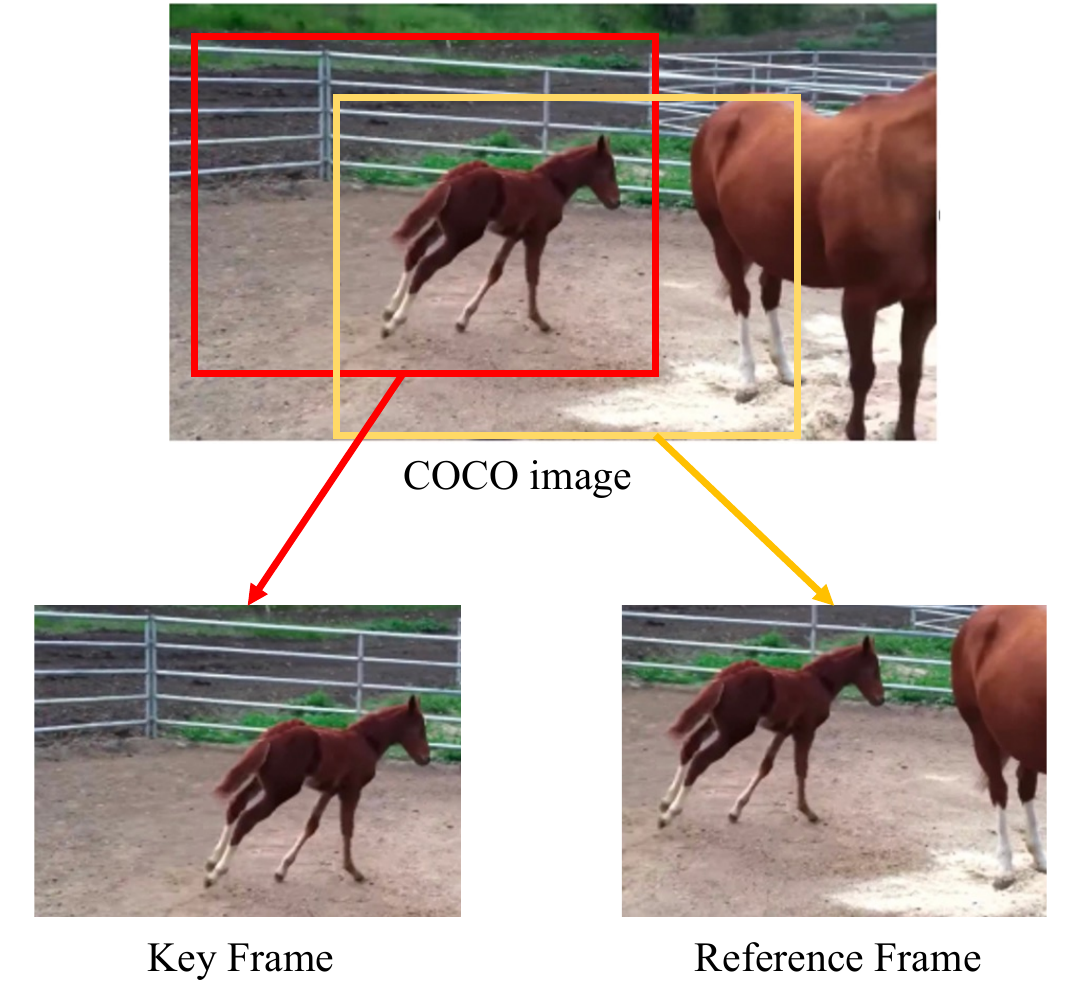}
\caption{Form pseudo key-reference frame pair by random crop.}
\label{fig:pseudo}
\vspace{-2ex}
\end{figure}

\subsection{Pseudo Key-reference Frame Pair}

Since the videos of this challenge have lower sampling rate, the gap in the appearance similarity of objects between the two frames is larger, which requires contrastive embeddings with higher temporal consistency.
To achieve this, we randomly and independently crop the image from COCO twice to form a pseudo key-reference frame pair, which is used to pre-train the contrastive embedding of our models.
As shown in Fig~\ref{fig:pseudo}, if we randomly crop twice on a coco image, we can get two sub-images on which the same object appears in different positions. 
They can be regarded as two adjacent key-reference frame pair with camera motion.
Pre-training IDOL on these pseudo frames requires that the embeddings belonging to the same horse are as close as possible in the embedding space. This forces the model to learn a position-insensitive contrastive embedding that relies on appearance of the object rather than the spatial position.

\section{Experiments}

\subsection{Implementation Details}
We use the same setting for Deformable DETR and the dynamic mask head following IDOL~\cite{IDOL}. 
The models are first pre-trained on the MS COCO 2017~\cite{coco} dataset for instance segmentation and then pre-trained on pseud key-reference frames from COCO.
Finally, the models are trained on YouTube-VIS 2022 training set for 12000 iterations, 
the learning rate is decayed by a factor of 0.1 at the 6000 iterations.
For data augmentation in training, we use multi-scale training scales at
[320, 352, 392, 416, 448, 480, 512, 544, 576, 608, 640] for shortest side.
During inference, the input frames are downscaled that the shortest side is at 480 pixels by default.
For multi-scale testing, the shortest side is at [480, 640, 800].
The model is trained on 8 A100 GPUs, with 4 pairs of frames per GPU.

\subsection{Main Results}

\begin{table}[h]
\centering
\resizebox{1.0\columnwidth}{!}{
\begin{tabular}{lccccc}
\toprule
Team  &$\rm mAP$ &$\rm AP_{50}$  &$\rm AP_{75}$ &$\rm AR_{1}$  &$\rm AR_{10}$  \\
\midrule
IIG    &\textbf{42.9}  &60.7  &\textbf{46.8}  &\textbf{35.0}  &51.4   \\
\textbf{Ours}  &40.2  &\textbf{61.1} &41.7  &32.6 &\textbf{55.0} \\
kyriemelon &38.0  &55.8   &39.2  &32.4  &48.0       \\
Starburst  &38.0  &59.9   &37.8   &33.5   &47.7  \\
Carl-Huang  &36.3  &56.8   &37.1   &32.2   &45.9 \\
\bottomrule
\end{tabular}
} 
\caption{Comparison with other methods on the ECCV 2022 Youtube-VIS test set.}
\label{ytvis22_test}
\end{table}
\vspace{-2mm}

We evaluate the performance of our method by participating in the ECCV 2022 YouTube-VIS Long Video Challenge.
As shown in Table~\ref{ytvis22_test}, our method achieves 40.2 AP on the test set and achieve second place.

\subsection{Ablation Study}

In this section we study how we achieve the final results as show in Table~\ref{ablation}. 
The baseline is with ResNet-50 backbone and single-scale testing.
Integrated with the Swin Transformer backbone\cite{SwinTransformer}, our method achieves a much higher AP of 48.4.
The pseudo key-reference frame pair pre-train improves the AP from 48.4 to 50.7.
After that, we utilize multi-scale testing for further boosting performance.
Different from image instance segmentation, the IoU computation is carried out in both spatial domain and temporal domain.
Multi-scale testing can further improve the result from 50.7 to 52.0.
Finally, by ensembling Swin-L and ConvNext-L\cite{convnext} in the same way, we achieve the 53.6 AP.
All the results are evaluate on ECCV 2022 Youtube-VIS validation set.

\begin{table}[h]
\centering
\resizebox{0.7\columnwidth}{!}{
\begin{tabular}{lcccc}
\toprule
Method    &$\rm mAP$  &$\rm \Delta mAP$  \\
\midrule
ResNet-50  &39.2  & -    \\
Swin-L  &48.4  &9.2  \\
+pseudo frame &50.7 &2.3    \\
+multi-scale  &52.0   &1.3     \\
+multi-model  &53.6 &1.6     \\
\bottomrule
\end{tabular}
} 
\caption{Ablation study on the ECCV 2022 Youtube-VIS validation set.}
\label{ablation}
\end{table}
\vspace{-2mm}

\section{Conclusions}
In this work, we adopt IDOL for the ECCV 2022 YouTube-VIS Long Video Challenge.
In addition, we use pseudo labels to further imporve contrastive learning, so as to obtain more temporally consistent instance embedding to improve tracking.
We believe the simplicity and effectiveness of our method shall benefit further research.

{\small
\bibliographystyle{ieee_fullname}
\bibliography{egbib}
}

\end{document}